\documentclass{article}

\usepackage{arxiv}

\usepackage[utf8]{inputenc} 
\usepackage[T1]{fontenc}    
\usepackage{hyperref}       
\usepackage{url}            
\usepackage{booktabs}       
\usepackage{amsfonts}       
\usepackage{nicefrac}       
\usepackage{microtype}      
\usepackage{lipsum}

\usepackage{hyperref}
\usepackage{cleveref}

\usepackage{graphicx}
 \usepackage{algorithm}
\usepackage{algorithmic}

\usepackage{calrsfs}
\DeclareMathAlphabet{\pazocal}{OMS}{zplm}{m}{n}
\newcommand{\La}{\pazocal{L}}
\newcommand{\Ea}{\pazocal{E}}
\newcommand{\Fa}{\pazocal{F}}
\newcommand{\Aa}{\pazocal{A}}
\newcommand{\Ra}{\pazocal{R}}
\newcommand{\Sa}{\pazocal{S}}
\newcommand{\Ga}{\pazocal{G}}

\newcommand{\Ta}{\pazocal{T}}

\newcommand{\Ca}{\pazocal{C}}
\newcommand{\Pa}{\pazocal{P}}
\newcommand{\Da}{\pazocal{D}}

\newcommand{\Ma}{\pazocal{M}}

\usepackage{amssymb,amsmath}

\title{Argumentation-based Agents that\\ Explain their Decisions}

\author{
  Mariela Morveli-Espinoza \\
  Graduate Program in Electrical and Computer Engineering (CPGEI),\\
 Federal University of Technology - Paran\'{a} (UTFPR),
 Curitiba - Brazil\\
  \texttt{morveli.espinoza@gmail.com} \\
   \And
   Ayslan Possebom \\
  Graduate Program in Electrical and Computer Engineering (CPGEI),\\
 Federal University of Technology - Paran\'{a} (UTFPR),
 Curitiba - Brazil\\
   \texttt{possebom@gmail.com} \\
   \And
   Cesar Augusto Tacla \\
   Graduate Program in Electrical and Computer Engineering (CPGEI),\\
 Federal University of Technology - Paran\'{a} (UTFPR),
 Curitiba - Brazil\\
 \texttt{tacla@utfpr.edu.br} \\
}

\newtheorem{defn}{Definition}

\begin{document}

\maketitle

\begin{abstract}
Explainable Artificial Intelligence (XAI) systems, including intelligent agents, must be able to explain their internal decisions, behaviours and reasoning that produce their choices to the humans (or other systems) with which they interact. In this paper, we focus on how an extended model of BDI (Beliefs-Desires-Intentions) agents can be able to generate explanations about their reasoning, specifically, about the goals he decides to commit to. Our proposal is based on argumentation theory, we use arguments to represent the reasons that lead an agent to make a decision and use argumentation semantics to determine acceptable arguments (reasons). We propose two types of explanations: the partial one and the complete one. We apply our proposal to a scenario of rescue robots.

\end{abstract}


\section{Introduction}

Explainability of intelligent agents has gained attention in recent years due to their growing utilization in human applications such as recommendation or coaching systems in domains as e-health, UAVs (Unmanned Aerial Vehicle), or smart environments. In these applications, the outcomes returned by the agent-based systems can be negatively affected due to the lack of clarity and explainability about their dynamics and rationality. Thus, if these systems would have explainability abilities, then their understanding, reliability, and acceptance could be enhanced.

The BDI model \cite{rao1995bdi} has become possibly the best-known and best-studied model of practical reasoning agents. In this model, agents decide which actions to perform in order to achieve their goals, which are selected during the deliberation process\footnote{Practical reasoning involves a deliberation process, to decide what states of affairs to achieve, and a means-ends reasoning, to decide how to achieve such states.}. This means that BDI agents are able to select the goals they are going to commit to -- which are called intentions -- from a set of desires; however, they are not endowed with explainability abilities. 

Consider a scenario of a natural disaster, where a set of robot agents wander an area in search of people needing help. When a person is seriously injured, he/she must be taken to the hospital; otherwise, he/she must be sent to a shelter. After the rescue work, the robots can be asked -- by another robot or by a human -- for an explanation of why a wounded person was sent to the shelter instead of taking him/her to the hospital, or why the robot decided to take to the hospital a person $x$ first, instead of taking another person $y$. In this case, it is clear that it is important to endow the agents with the ability of explain their decisions, that is, to explain how and why a certain desire became (or not) an intention. In the case of BDI agents, the path of this explanation is made up of only one step, which happens because in BDI agents there is only two stages in the intention formation process: desires and intentions. This means that there is a lack of a fine-grained analysis of this process, which may improve and enrich the informational quality of the explanations. 

An extended model for intention formation has been proposed by Castelfranchi and Paglieri \cite{castelfranchi2007role}, which was named the Belief-based Goal Processing model (let us denote it by BBGP model). The BBGP model has four stages: (i) activation (denoted $\mathsf{ac}$)\footnote{Hereafter, these notations are used for differentiate the stages and the statuses of goals.}, (ii) evaluation (denoted $\mathsf{ev}$), (iii) deliberation (denoted $\mathsf{de}$), and (iv) checking (denoted $\mathsf{ck}$). Consequently, four different statuses for a goal are defined: (i)active (=desire and denoted $\mathsf{ac}$), (ii) pursuable (denoted $\mathsf{pu}$), (iii) chosen (denoted $\mathsf{ch}$) and (iv) executive (=intention and denoted $\mathsf{ex}$). When a goal passes the activation (resp. evaluation, deliberation, checking) stage, it becomes active (resp. pursuable, chosen, executive). Unlike the BDI model, where desires and intentions are different mental states, Castelfranchi and Paglieri argue that desires and intentions may be considered as goals at different stages of processing.

Argumentation is a process of constructing and comparing arguments considering the conflicts -- which are called \hbox{attacks --} among them. The output of argumentation process is a set (or sets) of arguments -- called extensions -- which are internally consistent \cite{dung1995acceptability}. In the intention formation process, arguments can represent reasons for a goal to change (or not) its status. Thus, one can see the intention formation process as a decision-making process, where an agent has to decide which goal(s) passes a given stage and which does not. Adopting an argumentation-based approach in a decision making problem has some benefits. For example, a (human) user will obtain a ``good'' choice along with the reasons underlying this recommendation. Besides, argumentation-based decision-making is more similar with the way humans deliberate and finally make a choice \cite{ouerdane2010argumentation}. 

In \cite{espinoza2019argumentation}, the authors proposed an argumentation-based approach to computationally formalize the BBGP model. They used argumentation to support and justify the passage (change of the status) of the goals from their initial status until their final status. However, they did not endow BBGP-based agents with the ability of explaining the decisions about their goals. In this article, we aim to fill up this gap by equipping BBGP-based agents with a structure and a mechanism to generate such explanations.

Next section focuses on the knowledge representation. Section \ref{bbgp} presents the building blocks necessary for the generation of explanations. Section \ref{propo} presents our proposal for generating partial and complete explanations. Section \ref{aplicacion1} is devoted to the application of the proposal to the scenario of the rescue robots. Section \ref{relato} presents the main related work. Finally, Section \ref{conclus} is devoted to conclusions and future work.

\section{Preliminaries}

In this paper, BBGP-based agents use rule-based systems\footnote{These systems distinguish between facts, strict rules, and defeasible rules. A strict rule encodes strict information that has no exception, whereas a defeasible rule expresses general information that may have exceptions.
} as their basic reasoning model. The underlying logical language -- denoted by $\La$ -- consists of a set of literals\footnote{Literals are defined as positive or negative atoms where an atom is an n-ary predicate.} in first-order logical language. We represent non-ground formulae with Greek letters ($\varphi, \psi, ...$), variables with Roman letters ($x,y, ...$) and we name rules with $r_1, r_2, ...$. Strict rules are of the form $r=\varphi_1, . . . , \varphi_n \rightarrow \psi$, and defeasible rules are of the form $r=\varphi_1, . . . , \varphi_n \Rightarrow \psi$. Thus, a theory is a triple $\Ta=\langle \Fa, \Sa, \Da \rangle$ where $\Fa \subseteq \La$ is a set of facts, $\Sa$ is a set of strict rules, and $\Da$ is a set of defeasible rules. New information is produced from a given theory by applying the following concept, which was given in \cite{amgoud2013formal}.

\begin{defn}\label{def-der-sch} \textbf{(Derivation schema)} Let $\Ta=\langle \Fa, \Sa, \Da \rangle$ be a theory and $\psi \in \La$. A derivation schema for $\psi$ from $\Ta$ is a finite sequence $T = \{ (\varphi_1,r_1), ..., (\varphi_n, r_n)\}$ such that:

\begin{itemize}

\item $\varphi_n = \psi$
\item for $i=1 ... n$, $\varphi_i \in \Fa$ and $r_i=\emptyset$, or $r_i \in \Sa \cup \Da$ 
\end{itemize}
Based on a derivation scheme $T$ we also define the following sets:
$\mathtt{SEQ}(T) = \{\varphi_1, ..., \varphi_n\}$, $\mathtt{FACTS}(T) = \{\varphi_i |i \in \{1, ..., n\}, r_i=\emptyset\}$, $\mathtt{STRICT}(T)= \{r_i$ | $i \in \{1, ..., n\}, r_i \in \Sa\}$, $\mathtt{DEFE}(T)= \{r_i$ | $i \in \{1, ..., n\}, r_i \in \Da\}$.
\end{defn}

\section{The Building Blocks}
\label{bbgp}

In this section, we present the main structures necessary for the generation of partial and complete explanations.

From $\La$, we distinguish the following finite sets: 
(i) $\Fa$ is the set of facts of the agent and (ii) $\Ga$ is the set of goals of the agent. $\Fa$ and $\Ga$ are subsets of ground literals from the language $\La$ and are pairwise disjoint. Besides,  $\Ga=\Ga_\mathsf{ac} \cup \Ga_\mathsf{pu} \cup \Ga_\mathsf{ch} \cup \Ga_\mathsf{ex}$, where $\Ga_\mathsf{ac}$ (resp. $\Ga_\mathsf{pu}$, $\Ga_\mathsf{ch}$, $\Ga_\mathsf{ex}$) stands for the set of active (resp. pursuable, chosen, executive) goals. It holds that $\Ga_x \cap \Ga_y = \emptyset$, for $x,y \in \{\mathsf{ac, pu, ch, ex}\}$ with $x \neq y$.

Other important structures are the rules, which express the relation between the beliefs and the goals. The rules can be classified in standard and non-standard rules (activation, evaluation, deliberation, and checking rules). The former are made up of beliefs in both their premises and their conclusions and the latter are made up of beliefs in their premises and goals or beliefs about goals in their conclusions. Both standard and non-standard rules can be strict or defeasible rules. Standard rules can be used in any of the stages of the goal processing whereas non-standard rules are distinct for each stage. Thus, we have:

\noindent- \textbf{\textit{Standard rules}} ($r_{st})$: $\bigwedge \varphi_i \rightarrow\phi$ (or $\bigwedge \varphi_i \Rightarrow \phi$).\\
- \textbf{\textit{Activation rules}} ($r_{ac})$: $\bigwedge \varphi_i \rightarrow \psi$ (or $\bigwedge \varphi_i \Rightarrow \psi$). \\
- \textbf{\textit{Evaluation rules}} ($r_{ev}$): $\bigwedge\varphi_i \rightarrow \neg \psi$ (or $\bigwedge\varphi_i \Rightarrow \neg \psi$). \\
- \textbf{\textit{Deliberation rules}}: $r_{de}^1: \neg has\_incompatibility(g) \rightarrow chosen(g) $\\
\hspace*{3cm}$r_{de}^2: most\_valuable(g) \rightarrow chosen(g) $\\
- \textbf{\textit{Checking rule}}: $r_{ck}= has\_plans\_for(g)\wedge$ $satisfied\_context\_for(g) \rightarrow executive(g) $

Where $\varphi_i$ and $\psi$ denote non-ground literals that represent beliefs and goals, respectively.  $g$ denote ground literals that represent a goal\footnote{In any of the states of the goal processing, a goal is represented by a ground atom. However, before a goal becomes active, it has the form of a non-ground atom; in this case, we call it a sleeping goal. Thus, $\psi$ is a sleeping goal and $g$ a goal in some status.}. Notice that standard, activation, and evaluation rules are designed and entered by the programmer of the agent, and their content is dependent on the application domain. Otherwise, rules in
deliberation and checking stages are pre-defined and no new rules of these types can be defined
by the user. Finally, let $\Ra_{st}, \Ra_{ac}, \Ra_{ev}, \Ra_{de}$, and $\Ra_{ck}$ denote the set of standard, activation, evaluation, deliberation, and checking rules, respectively. Next, we define the theory of a BBGP-based agent. 

\begin{defn} \textbf{(BBGP-based Agent Theory)} A theory is a triple $\Ta=\langle \Fa,\Sa, \Da \rangle$ such that: (i) $\Fa$ is the set of beliefs of the agent, (ii) $\Sa=\Ra_{st}' \cup \Ra_{ac}' \cup \Ra_{ev}' \cup \Ra_{de}' \cup \Ra_{ck}'$ is the set of strict rules, and (ii)  $\Da=\Ra_{st}'' \cup \Ra_{ac}'' \cup \Ra_{ev}'' \cup \Ra_{de}'' \cup \Ra_{ck}''$ is the set of defeasible rules, where $\Ra_{x}=\Ra_{x}' \cup \Ra_{x}''$ (for $x \in \{st,ac,ev,de,ck\}$). It holds that $\Ra_{x}' \cap \Ra_{x}'' =\emptyset$.

\end{defn}

From a theory, a BBGP-based agent can build arguments. There are two categories of arguments. The first one -- called epistemic arguments -- justifies or attacks beliefs, while the other one -- called stage arguments -- justifies or attacks the passage of a goal from one stage to another. There is a set of arguments for each stage of the BBGP model.

\begin{defn}\textbf{(Arguments)} Let $\Ta=\langle \Fa, \Sa, \Da \rangle$ be a BBGP-based agent theory and $\Ta'=\langle \Fa, \Ra_{st}', \Ra_{st}'' \rangle$ and $\Ta''=\langle \Fa, \Sa', \Da' \rangle$ be two sub-theories of $\Ta$, where $\Sa' = \Sa \setminus \Ra_{st}'$ and $\Da' = \Da \setminus \Ra_{st}''$. An \textit{\textbf{epistemic argument}} constructed from $\Ta'$ is a pair $A=\langle T, \varphi \rangle$ such that:

(1) $\varphi \in \La$\\
\indent(2) $T$ is a derivation schema for $\varphi$ from $\Ta'$

\noindent On the other hand, a \textit{\textbf{stage argument}} constructed from $\Ta''$ is a pair $A= \langle T, g \rangle$ such that: 

(1) $g \in \Ga$ \\
\indent(2) For the activation and evaluation stages: $T$ is a derivation schema for $g$ from $\Ta''$. For the deliberation stage: $T$ is a derivation schema for $chosen(g)$ from $\Ta''$. For the checking stage: $T$ is a derivation schema for $executive(g)$ from $\Ta''$.

For both kinds of arguments, it holds that $\mathtt{SEQ}(T)$ is consistent\footnote{A set $\La' \subseteq \La$ is consistent iff $\nexists \varphi, \varphi' \in \La'$ such that $\varphi=\neg \varphi'$. It is inconsistent otherwise.} and $T$ must be minimal\footnote{Minimal means that there is no $T' \subset T$ such that $\varphi$ ($g$, $chosen(g),$ or $executive(g)$) is a derivation schema of $T'$.}. Finally, 
$\mathtt{ARG}_{ep}$, $\mathtt{ARG}_{ac}$, $\mathtt{ARG}_{ev}$, $\mathtt{ARG}_{de}$, and $\mathtt{ARG}_{ck}$ denote the set of all epistemic, activation, evaluation, deliberation, and checking arguments, respectively. As for notation, $\mathtt{CLAIM}(A)=\varphi$ (or $g$) and $\mathtt{SUPPORT}(A)=T$ denote the conclusion and the support of an argument $A$, respectively.

\end{defn}

\vspace*{-0.15cm}

An argument may have a set of sub-arguments. Thus, an argument $\langle T', \varphi'\rangle$ is a \textit{\textbf{sub-argument}} of $\langle T, \varphi \rangle$ iff $\mathtt{FACTS}(T') \subseteq \mathtt{FACTS}(T), \mathtt{STRICT}(T')\subseteq  \mathtt{STRICT}(T),$ and $\mathtt{DEFE}(T') \subseteq \mathtt{DEFE}(T)$. $\mathtt{SUB}(A)$ denotes the set of all sub-arguments of $A$.

Stage arguments built from $\Ta$ constitute a cause for a goal changes its status. However, it is not a proof that the the goal should adopt another status. The reason is that an argument can be attacked by other arguments. Two kinds of attacks are distinguished: (i) the attacks between epistemic arguments, and (ii) the mixed attacks, in which an epistemic argument attacks a stage argument. The former is defined over $\mathtt{ARG}_{ep}$ and is captured by the binary relation $\mathsf{att}_{ep}\: \subseteq \mathtt{ARG}_{ep} \times \mathtt{ARG}_{ep}$. The latter is defined over $\mathtt{ARG}_{ep}$ and $\mathtt{ARG}_{x}$ (for $x\in \{ac,ev,de,ck\}$); and is captured by the binary relation $\mathsf{att}_{mx}\: \subseteq \mathtt{ARG}_{ep} \times \mathtt{ARG}_x$. For both kinds of attacks, $(A, B) \in \mathsf{att}_{mx}$ (or $(A, B) \in \mathsf{att}_{ep}$)  denotes that there is an attack relation between arguments $A$ and $B$. Next definition captures both kinds of attacks; thus, rebuttal may occur only between epistemic arguments and undercut may occur in both kinds of attacks.


\begin{defn} \textbf{(Attacks)} Let $\langle T',\varphi' \rangle$ and $\langle T,\varphi \rangle$ be two epistemic arguments, and $\langle T,g \rangle$ be a stage argument. $\langle T',\varphi' \rangle$ \textit{\textbf{rebuts}} $\langle T,\varphi \rangle$ if $\varphi = \neg \varphi'$. $\langle T,\varphi \rangle$ \textit{\textbf{undercuts}} $\langle T',\neg \varphi' \rangle$ (or $\langle T,g \rangle$) if $\varphi' \in \mathtt{FACTS}(T)$.

\end{defn}

From epistemic and stage arguments and the attacks between them, it is generated a different Argumentation Framework (AF) for each stage of the BBGP model.

\begin{defn} \textbf{(Argumentation Framework)} \label{defargfra} An argumentation framework $\Aa\Fa_x$ is a pair $\Aa\Fa_x=\langle \mathtt{ARG},\mathsf{att} \rangle$ ($x \in \{ac,ev,de,ck\}$) such that:
\begin{itemize}
\item $\mathtt{ARG}= \mathtt{ARG}_x \cup \mathtt{ARG}_{ep}' \cup \mathtt{SUBARGS}$ , where $\mathtt{ARG}_x$ is a set of stage arguments, $\mathtt{ARG}_{ep}'=\{A\: |\: A \in \mathtt{ARG}_{ep}$ and $(A,B) \in \mathsf{att}_{mx}$ or $(A,C) \in \mathsf{att}_{ep}\}$, where $B \in \mathtt{ARG}_x$ and $C \in \mathtt{ARG}_{ep}'$, and $\mathtt{SUBARGS}= \bigcup_{A \in \mathtt{ARG}_x, A \in \mathtt{ARG}_{ep}'} \mathtt{SUB}(A)$.

\item $\mathsf{att} = \mathsf{att}_{ep}' \cup \mathsf{att}_{mx}'$, where $\mathsf{att}_{ep}' \subseteq \mathtt{ARG}_{ep}' \times \mathtt{ARG}_{ep}'$ and $\mathsf{att}_{mx}' \subseteq \mathtt{ARG}_{ep}' \times \mathtt{ARG}_x$.
 \end{itemize}
\end{defn}

The next step is to evaluate the arguments that make part of the AF. This evaluation is important because it determines which goals pass (acceptable goals) from one stage to the next. The aim is to obtain a subset of $\mathtt{ARG}$ without conflicting arguments. In order to obtain it, we use an \textit{acceptability semantics}, which return one or more sets -- called extensions -- of acceptable arguments. The fact that a stage argument belong to an extension determines the change of status of the goal in its claim. Next, the main semantics introduced by Dung \cite{dung1995acceptability} are recalled\footnote{It is not the scope of this article to study the most adequate semantics for goal processing or the way to select an extension when more than one is returned by a semantics. For a brief study of these issues, the reader is referred to \cite{espinoza2019argumentation}.}.

\begin{defn} \textbf{(Semantics)} Let $\Aa\Fa_x=\langle \mathtt{ARG},\mathsf{att} \rangle$ be an AF (with $x \in \{ac,ev,de,ck\}$) and $\Ea \subseteq \mathtt{ARG}$:

\noindent- $\Ea$ is \textbf{\textit{conflict-free}} if $\forall A,B \in \Ea$, $(A,B) \notin \mathsf{att}$ or $(B,A) \notin \mathsf{att}$\\
- $\Ea$ \textbf{\textit{defends}} $A$ iff $\forall B \in \mathtt{ARG}$, if $(B, A) \in \mathsf{att}$, then $ \exists C \in \Ea$ s.t. $(C,B) \in \mathsf{att}$.\\
- $\Ea$ is \textbf{\textit{admissible}} iff it is conflict-free and defends all its elements. \\
- A conflict-free $\Ea$ is a \textbf{\textit{complete extension}} iff we have $\Ea=\{A | \Ea$ defends $A\}$. \\
- $\Ea$ is a \textbf{\textit{preferred extension}} iff it is a maximal (w.r.t the set inclusion) complete extension.\\
- $\Ea$ is a \textbf{\textit{grounded extension}} iff is a minimal (w.r.t. set inclusion) complete extension.\\
- $\Ea$ is a \textbf{\textit{stable extension}} iff $\Ea$ is conflict-free and $\forall A \in \mathtt{ARG}$, $\exists B \in \Ea$ such that $(B,A) \in \mathtt{att}$.

\end{defn}

\section{Partial and Complete Explanations}
\label{propo}

In order to able to generate partial and complete explanations, a BBGP-based agent needs to store information about the progress of his goals, that is, the changes of the statuses of such goals and the causes of these changes. The latter are stored in each AF in form of accepted arguments; however, the former cannot be stored in an AF. Thus, we need a structure to saves both the status of each goal and the AF that supports this status. Considering that at each stage, the agent generates arguments and attacks for more than one goal -- which are stored in each $\Aa\Fa_x$ -- and we only need those arguments and attacks related to one goal, we have to extract from $\Aa\Fa_x$ such arguments and attacks. In other words, we need to obtain a sub-AF.

\begin{defn} \textbf{(Sub-AF)} Let $\Aa\Fa_x=\langle \mathtt{ARG},\mathsf{att} \rangle$ (with $x \in \{ac,ev, de,ck\}$) be the an AF and $g \in \Ga$ a goal. An AF $\Aa\Fa_x'=\langle \mathtt{ARG}', \mathsf{att}' \rangle$ is a \textit{sub-AF} of $\Aa\Fa_x$ (denoted $\Aa\Fa_x' \sqsubseteq \Aa\Fa_x$), if $\mathtt{ARG}' \subseteq \mathtt{ARG}$ and $\mathsf{att}'$ $=$ $\mathsf{att} \otimes \mathtt{ARG}'$, s.t.:

\noindent- $\mathtt{ARG}' =\{\{A | A \in \mathtt{ARG}_x, \mathtt{CLAIM}(A)=g\} \cup \{B | (B,A) \in \mathsf{att}_{mx}$ or $(B,C) \in \mathsf{att}_{ep}'\}$, where $B \in \mathtt{ARG}_{ep}$, $C \in \mathtt{ARG}_{ep}',  \mathsf{att}_{ep}' \subset \mathsf{att}',$ and $\mathtt{ARG}_{ep}' \subset \mathtt{ARG}'\}\}$, and \\
- $\mathsf{att} \otimes \mathtt{ARG}'$ returns a subset of $\mathsf{att}$ that involves just the arguments in $\mathtt{ARG}'$. 

\end{defn}

Next, we define the structure that stores the causes of the change of the status of a goal, which must be updated after a new change occurs. We can see this structure as a table record, where each row saves the status of a goal along with the AF that supports such status.

\begin{defn}\textbf{(Goal Memory)} Let $\Aa\Fa_x=\langle \mathtt{ARG},\mathsf{att} \rangle$ be an AF (with $x \in \{ac,ev,de,ck\}$), $\Aa\Fa_x' \sqsubseteq \Aa\Fa_x$ a sub-AF, and $g \in \Ga$ a goal. The goal memory $\Ga\Ma_g$ for goal $g$ is a set of ordered pairs $(\mathtt{STA},\mathtt{REASON})$ such that:

\begin{itemize}
\item $\mathtt{STA} \in \{\mathsf{ac, pu, ch,ex,} \mathsf{not}\; \mathsf{ac}, \mathsf{not}\; \mathsf{pu}, \mathsf{not}\; \mathsf{ch}, \mathsf{not}\; \mathsf{ex}\}$ where $\{\mathsf{ac, pu, ch,ex}\}$ represent the status $g$ attains due to the arguments in $\mathtt{REASON}$ whereas $\{\mathsf{not}\; \mathsf{ac}, \mathsf{not}\; \mathsf{pu}, \mathsf{not}\; \mathsf{ch}, \mathsf{not}\; \mathsf{ex}\}$ represent the status $g$ cannot attain due to the arguments in $\mathtt{REASON}$.
\item $\mathtt{REASON}=\Aa\Fa_x' \sqsubseteq \Aa\Fa_x$ is a sub-AF whose selected extension supports the current status of $g$. 

\end{itemize}
\end{defn}

Let $\Ga\Ma^+$ be the set of all goal memories and $\mathtt{NUM\_REC}: \Ga\Ma^+ \rightarrow \mathbb{N}$ a function that returns the number of records of a given $\Ga\Ma$. From the goal memory structure, the partial and complete explanation can be generated.

\begin{defn}\textbf{(Partial and Complete Explanations)}  Let $g \in \Ga$ be a goal, $\Ga\Ma_g$ the memory of $g$, and $\Aa\Fa_{ac}$, $\Aa\Fa_{ev}$, $\Aa\Fa_{de}$, and $\Aa\Fa_{ck}$ the four argumentation frameworks involved in the goal processing. Besides, let $x \in \{\mathsf{ac,pu,ch,ex}\}$ denote the current status of $g$:

\begin{itemize}

\item A \textbf{complete explanation} $\Ca\Ea_g$ for $g \in \Ga_x$ is obtained as follows: $\Ca\Ea_g= \bigcup_{i=1}^{i=\mathtt{NUM\_REC}(\Ga\Ma_g)} \mathtt{REASON}_i$, where $\mathtt{REASON}_i \sqsubseteq \Aa\Fa_{ac}$, $\mathtt{REASON}_i \sqsubseteq \Aa\Fa_{ev}$, $\mathtt{REASON}_i \sqsubseteq \Aa\Fa_{de}$, and $\mathtt{REASON}_i \sqsubseteq \Aa\Fa_{ck}$.

\item A \textbf{partial explanation} $\Pa\Ea_g$ is obtained as follows: $\Pa\Ea_g= \bigcup_{i=1}^{i=\mathtt{NUM\_REC}(\Ga\Ma_g)} \Ea_i$, where $\Ea_i$ is the selected  extension obtained from $\mathtt{REASON}_i$.

\end{itemize}
\end{defn}

\section{Application: Rescue robots scenario}
\label{aplicacion1}

In this section, we present the application of the proposed approach to the rescue robots scenario.

Firstly, let us present the mental states of the robot agent, let us call him $\mathtt{BOB}$:

$\Ga_\mathsf{sl}=\{\mathsf{g_\mathsf{sl}^1,g_\mathsf{sl}^2,g_\mathsf{sl}^3}\}$, $\Ga=\{\}$, which means that $\Ga_\mathsf{ac}=\{\}$,\\
\indent $\Ga_\mathsf{pu}=\{\}$, $\Ga_\mathsf{ch}=\{\}$, and $\Ga_\mathsf{ex}=\{\}$\\
\indent$\Ra_{st}=\{\mathsf{r_{st}^1,r_{st}^2,r_{st}^3,r_{st}^4}\}$, $\Ra_{ac}=\{\mathsf{r_{ac}^1,r_{ac}^2,r_{ac}^3}\}$,\\
\indent $\Ra_{ev}=\{\mathsf{r_{ev}^1,r_{ev}^2}\}$, $\Ra_{de}=\{\mathsf{r_{de}^1,r_{de}^2}\}$, and $\Ra_{ck}=\{\mathsf{r_{ck}}\}$\\
\indent$\Fa=\{\mathsf{b_1,b_2, b_3, b_4,b_5, \neg b_6, b_8,b_9,b_{10}, b_{11},b_{12},b_{13}}\}$  

The detail of each set is presented below:

\begin{table}[!h]
\begin{tabular}{l}
\hline
Sleeping goals\\
\hline
- $\mathsf{g_s^1=take\_hospital(x})$ \textit{ //take a person x to the hospital} \\ 
- $\mathsf{g_s^2=go(x,y)}$ \textit{//go to zone (x,y)}\\
- $\mathsf{g_s^3=send\_shelter(x)}$ \textit{//send a person x to the shelter}\\
\hline
\hline
Standard Rules\\
\hline
- $\mathsf{r_{st}^1=new\_supply(x) \Rightarrow available(x)}$  \textit{//if there is a new supply x,} \textit{ then x is available}\\
- $\mathsf{r_{st}^2=has\_fract\_bone(x) \Rightarrow injured\_severe(x)}$  \textit{//if x has a fractured}
 \textit{bone, then x is severely injured}\\
- $\mathsf{r_{st}^3=fract\_bone(x,arm)  \Rightarrow \neg injured\_severe(x)} $  \textit{//if the fractured bone}
 \textit{is in the arm, then x is not severely injured}\\
- $\mathsf{r_{st}^4=open\_fracture(x) \rightarrow injured\_severe(x)}$  \textit{//if x has an open fracture, }
\textit{then x is severely injured}\\
\hline
\hline
Activation rules\\
\hline
- $\mathsf{r_{ac}^1=injured\_severe(x) \Rightarrow take\_hospital(x)}$ \textit{//if person $x$ is severely}
 \textit{injured, then take $x$ to the hospital}\\
- $\mathsf{r_{ac}^2= \neg injured\_severe(x) \Rightarrow send\_shelter(x)}$ \textit{//if person $x$ is not}
\textit{severely injured, then send $x$ to the shelter}\\
- $\mathsf{r_{ac}^3=asked\_for\_help(x,y) \Rightarrow go(x,y)}$  \textit{//if $\mathtt{BOB}$ is asked for help in}
\textit{zone $(x,y)$, then go to that zone}\\
\hline
\hline

Evaluation rules\\
\hline
- $\mathsf{r_{ev}^1=greater(weight(x),80) \rightarrow \neg take\_hospital(x)}$  \textit{//If person $x$ weights more than 80 kilos, then it is not }\\
\textit{possible to take him/her to the hospital}\\
- $\mathsf{r_{ev}^2=\neg available(bed,x) \Rightarrow \neg take\_hospital(x)}$  \textit{//If there is no available  bed for x, then it is not possible to}\\
\textit{ take $x$ to hospital}\\
\hline
\hline

\end{tabular}
\end{table}

\begin{table}[!h]

\begin{tabular}{l}
\hline
\hline
Beliefs\\
\hline
- $\mathsf{b_1=be\_operative(me)}$\\
- $\mathsf{b_2=has\_fract\_bone(man\_32)}$ \textit{//There is a 32-year-old man with a} \textit{fractured bone} \\
- $\mathsf{b_3=fract\_bone(man\_32,arm)}$ \textit{//The 32-year-old man has a fractured }\textit{arm.}\\
- $\mathsf{b_4=asked\_for\_help(2,6)}$ \textit{//There is an aid request in slot (2,6).}\\
- $\mathsf{b_5=open\_fracture(man\_32)}$ \textit{//The 32-year-old man has an open}\textit{ fracture.}\\
- $\mathsf{\neg b_6= \neg available(bed, man\_32)}$ \textit{//There is no an available bed.}\\
- $\mathsf{ b_8= new\_supply(bed)}$ \textit{//There is a new supply.}\\
- $\mathsf{ b_9= weight(man\_32,70)}$ \textit{//$man\_32$ weights 70 kg.}\\
- $\mathsf{ b_{10}= has\_plans\_for('take\_hospital(man\_32)')}$ \\
- $\mathsf{ b_{11}= has\_plans\_for('go(2,6)')}$ \\
- $\mathsf{ b_{12}=most\_valuable\_goal('take\_hospital(man\_32)')}$\\
- $\mathsf{ b_{13}= satisfied\_context\_for('take\_hospital(man\_32)')}$\\
\hline

\end{tabular} 
\end{table}

Thus, the theory of agent $\mathtt{BOB}$ is: $\Ta=\langle \Fa, \Sa, \Da \rangle$ where $\Fa=\{\mathsf{b_1,b_2, b_3, b_4,b_5, \neg b_6,b_8,b_9,} \mathsf{b_{10},b_{11},b_{12},b_{13}}\}$, $\Sa=\{\mathsf{r_{ev}^1,r_{de}^1,r_{de}^2,r_{ck}, r_{st}^4,}\}$, and $\Da=\{\mathsf{r_{st}^1,  r_{ac}^1, r_{ac}^2,  r_{ev}^2,r_{st}^2, r_{st}^3, r_{ac}^3}\}$.

\subsection{Generated Argumentation Frameworks}

Based on the mental state of agent $\mathtt{BOB}$, the arguments and AFs for each stage can be generated. For this application, we will use preferred semantics for calculating the extensions.

For the \textbf{activation stage} seven epistemic arguments and four activation arguments can be built:\\
\indent$\mathsf{A_{ep}^2=\langle \{(b_2,\emptyset)\},b_2\rangle},$ $\mathsf{A_{ep}^3=\langle \{(b_3,\emptyset)\},b_3\rangle}$\\
\indent$\mathsf{A_{ep}^4=\langle \{(b_4,\emptyset)\},b_4\rangle}$, $\mathsf{A_{ep}^5=\langle \{(b_5,\emptyset)\},b_5\rangle}$\\
\indent$\mathsf{A_{ep}^7=\langle \{(b_2,\emptyset),(b_7,r_{st}^2)\},b_7\rangle}$,$\mathsf{A_{ep}^8= \langle \{(b_3,\emptyset),(\neg b_7,r_{st}^3) \},\neg b_7\rangle}$\\
\indent$\mathsf{A_{ep}^9= \langle \{(b_5,\emptyset),(b_7,r_{st}^4) \},b_7\rangle}$,$\mathsf{A_{ac}^1= \langle \{(b_4,\emptyset),(g_1,r_{ac}^3) \}, g_1\rangle}$\\
\indent$\mathsf{A_{ac}^2= \langle \{(b_2,\emptyset),(b_7,r_{st}^2),}$, $\mathsf{A_{ac}^3= \langle \{(b_5,\emptyset),(b_7, r_{st}^4),\rangle}$\\
\indent$\mathsf{A_{ac}^4= \langle \{(b_3,\emptyset),(\neg b_7, r_{st}^3),\rangle}$

The AF for this stage is: $\Aa\Fa_{ac}=\langle \mathsf{ \{A_{ep}^1, A_{ep}^2,A_{ep}^3,A_{ep}^4, A_{ep}^5,}\mathsf{A_{ep}^7,A_{ac}^1,A_{ac}^2,A_{ac}^3,A_{ac}^4\},}   \mathsf{ \{ (A_{ep}^7, A_{ep}^8),(A_{ep}^8, A_{ep}^7), (A_{ep}^8, A_{ep}^9),}\\ \mathsf{(A_{ep}^9, A_{ep}^8),(A_{ep}^7, A_{ac}^4),(A_{ep}^8, A_{ac}^3),(A_{ep}^8, A_{ac}^2),(A_{ep}^9, A_{ac}^4)} \}\rangle$. 
The next step is to evaluate the acceptability of the arguments. We first apply the preferred semantics to $\Aa\Fa_{ac}$, the result is: $\Ea=\{\mathsf{A_{ep}^2,A_{ep}^3, A_{ep}^4,A_{ep}^5,A_{ep}^7,A_{ep}^9,A_{ac}^1,A_{ac}^2, A_{ac}^3\} }$. Therefore, we have that the set of justified conclusions is: $\{\mathsf{b_2,b_3,b_4,}\break \mathsf{b_5,b_7,g_1,g_2}\}$. Thus, the set of justified goals is $\{\mathsf{g_1, g_2}\}$. This means that robot agent $\mathtt{BOB}$ activates goals $\mathsf{g_1=go(2,6)}$ and $\mathsf{g_2=take\_hospital(man\_32)}$ but he does not activate goal $\mathsf{g_3=send\_shelter(man\_32)}$. Therefore, we have that $\Ga_\mathsf{ac}=\{\mathsf{g_1,g_2} \}, \Ga_\mathsf{pu}=\{\}, \Ga_\mathsf{ch}=\{\},$ and $ \Ga_\mathsf{ex}=\{\}$.

For the \textbf{evaluation stage}, the epistemic and evaluation arguments that can be built from the mental state of $\mathtt{BOB}$ are:\\
\indent$\mathsf{A_{ep}^6=\langle \{(\neg b_6,\emptyset)\},\neg b_6\rangle}$, $\mathsf{A_{ep}^{10}=\langle \{ (b_8,\emptyset)\}, b_8\rangle}$ \\ 
\indent$\mathsf{A_{ep}^{11}= \langle \{(b_8,\emptyset),(b_6,r_{st}^1) \}, b_6\rangle}$  \\ 
\indent $\mathsf{A_{ev}^1=\langle \{(\neg b_6,\emptyset),(\neg g_2,r_{ev}^2)\}, \neg g_2\rangle}$  

The AF for this stage is $\Aa\Fa_{ev}=\langle\{\mathsf{A_{ep}^6,A_{ep}^{10},A_{ep}^{11},A_{ev}^1\},}$ $\mathsf{\{(A_{ep}^6, A_{ep}^{11}),(A_{ep}^{11}, A_{ep}^6),} \mathsf{(A_{ep}^{11}, A_{ev}^1})\}\rangle$. We have two preferred extensions for $\Aa\Fa_{ev}$: $\Ea=\{\mathsf{\{A_{ep}^{10},A_{ep}^{11}\}, \{A_{ep}^6, A_{ep}^{10}, A_{ev}^1\}} \}$. Since the second preferred extension refrains a goal of becoming pursuable, the agent chooses the first preferred extension. Since there is no evaluation argument that belongs to the selected extension, we can say that the passage of both currently active goals is justified. Therefore, both $\mathsf{g_1= go(2,6)}$ and $\mathsf{g_2=take\_hospital(man\_32)}$ are now pursuable goals.


Regarding \textbf{deliberation stage}, the arguments generated for this stage are:\\
\indent$\mathsf{A_{ep}^{12}=\langle \{(b_{12},\emptyset)\}, b_{12}\rangle}$ \\
\indent$\mathsf{A_{de}^1=\langle \{(b_{12},\emptyset),(chosen(g_2),r_{de}^2)\}, g_2\rangle}$ 

The AF for this stage is $\Aa\Fa_{de}=\langle\{\mathsf{A_{ep}^{12},A_{de}^1}\},\{\}\rangle$. In this case, there is no attacks between the arguments and we have only one preferred extension for $\Aa\Fa_{de}$: $\Ea=\{\mathsf{A_{ep}^{12},A_{de}^1} \}$. We have that a deliberation argument belongs to the extension, so we can say that the passage of goal $\mathsf{g_2}$ is justified. Therefore, $\mathsf{g_2=take\_hospital(man\_32)}$ is now a chosen goal.


Finally, for the \textbf{checking stage}, the generated arguments are:
\indent$\mathsf{A_{ep}^{13}=\langle \{(b_{10},\emptyset)\}, b_{10}\rangle}$, $\mathsf{A_{ep}^{11}=\langle \{(b_{13},\emptyset)\}, b_{13}\rangle}$ \\
\indent$\mathsf{A_{ch}^1=\langle \{(b_{10},\emptyset),(b_{13},\emptyset),}\mathsf{(executive(g_2),r_{ch})\}, g_2\rangle}$

The AF for this stage is $\Aa\Fa_{ck}=\langle\{\mathsf{A_{ep}^{13},A_{ep}^{14},A_{ch}^1}\},\{\}\rangle$. There is only one preferred extension for $\Aa\Fa_{ck}$: $\Ea=\{\mathsf{A_{ep}^{13},A_{ep}^{14},A_{ch}^1} \}$. We have that a checking argument belongs to the extension, so we can say that the passage of goal $\mathsf{g_2}$ is justified. Therefore, $\mathsf{g_2=take\_hospital(man\_32)}$ is now an executive goal.

At last, we present the final configuration of $\Ga$: $\Ga_\mathsf{ac}=\{\}$, $\Ga_\mathsf{pu}=\{\mathsf{g_1}\}, \Ga_\mathsf{ch}=\{\},$ and $ \Ga_\mathsf{ex}=\{g_2\}$.

\subsection{Partial and Complete Explanations}

First of all, let us present the goal memories of goals $\mathsf{g_1}$, $\mathsf{g_2}$, and $\mathsf{g_3}$. Next table shows the sub-AFs that allow goal $\mathsf{g_1}$ to become pursuable, it also shows it cannot become chosen because there is no deliberation argument that supports its change of status.

\begin{table}[!h]
\label{tab-mrg1}
\begin{center}
\renewcommand{\arraystretch}{1.2}
\begin{tabular}{|c|l|}
\hline 
$\mathtt{STA}$ & $\mathtt{REASON}$ \\ 
\hline 
$ac$ & $\Aa\Fa_{ac}^{g_1}=\langle \{\mathsf{A_{ac}^1, A_{ac}^4}\},\{\}\rangle$ \\ 
$pu$ & $\Aa\Fa_{ev}^{g_1}=\langle \{\},\{\}\rangle$ \\ 
$\mathtt{not}$ $ch$ & $\Aa\Fa_{de}^{g_1}=\langle \{\},\{\}\rangle$ \\ 
\hline 
\end{tabular} 
\end{center}
\end{table}

In the following table, we have all the sub-AFs that allow goal $\mathtt{g_2}$ to become executive. Notice that $\mathsf{g_1}$ becomes pursuable because no evaluation argument refrains its passage to the deliberation stage; on the contrary, in sub-AF $\Aa\Fa_{ev}^{g_2}$ there is an evaluation argument against $\mathsf{g_2}$, which is attacked by an epistemic argument. Thus, $\mathsf{g_2}$ becomes pursuable due to the defence of one of the epistemic arguments of the sub-AF.

\begin{table}[!h]
\label{tab-mrg2}
\begin{center}
\renewcommand{\arraystretch}{1.2}
\begin{tabular}{|c|l|}
\hline 
$\mathtt{STA}$ & $\mathtt{REASON}$ \\ 
\hline 
$ac$ & $\Aa\Fa_{ac}^{g_2}=\langle \{\mathsf{A_{ac}^2, A_{ac}^3, A_{ep}^7,A_{ep}^2, A_{ep}^3, A_{ep}^9,A_{ep}^5, A_{ep}^8\},}$ \\ 
&\hspace*{1.3cm}$\{\mathsf{(A_{ep}^8,A_{ac}^2),(A_{ep}^7, A_{ep}^8),(A_{ep}^8, A_{ep}^7),(A_{ep}^9, A_{ep}^8)}\}\rangle$\\
$pu$ & $\Aa\Fa_{ev}^{g_2}=\langle \{\mathsf{A_{ev}^1,A_{ep}^{11},A_{ep}^6}\},\{\mathsf{(A_{ep}^{11},A_{ev}^1),(A_{ep}^{11},A_{ep}^6), (A_{ep}^6,A_{ep}^{11})}\}\rangle$ \\ 
$ch$& $\Aa\Fa_{de}^{g_2}=\langle \{ \mathsf{A_{de}^1, A_{ep}^{12}}\},\{\}\rangle$\\
$ex$& $\Aa\Fa_{ck}^{g_2}=\langle \{ \mathsf{A_{ck}^1, A_{ep}^{13},A_{ep}^{11}}\},\{\}\rangle$\\
\hline 
\end{tabular} 
\end{center}
\end{table}

Finally, next table shows that goal $\mathsf{g_3}$ cannot become active. Even though there is an activation argument, it is attacked by two epistemic arguments. Thus, after applying the semantics the activation argument is not part of the preferred extension.

\begin{table}[!h]
\label{tab-mrg3}
\begin{center}
\renewcommand{\arraystretch}{1.2}
\begin{tabular}{|c|l|}
\hline 
$\mathtt{STA}$ & $\mathtt{REASON}$ \\ 
\hline 
$\mathsf{not}$ $ac$ & $\Aa\Fa_{ac}^{g_3}=\langle \{ \mathsf{A_{ac}^4, A_{ap}^8, A_{ap}^2,A_{ap}^3,A_{ap}^5, A_{ap}^7, A_{ap}^9\},}$\\
&\hspace*{1,3cm}$\{(\mathsf{A_{ap}^7,A_{ac}^4),(A_{ap}^9, A_{ac}^4), (A_{ap}^8,A_{ap}^7), (A_{ap}^7,A_{ap}^8),}$ \\ 
&\hspace*{1,3cm}$\mathsf{(A_{ap}^9,A_{ap}^8)}\}\rangle$\\

\hline 
\end{tabular} 
\end{center}
\end{table}

 For the sake of simplicity, suppose that rescue robots can communicate with humans by means of natural language. Now, suppose that at the end of a rescue day, $\mathtt{BOB}$ is interrogated with the following question: \textit{Why have you taken to the hospital $\mathsf{man\_32}$ instead of sending him to the shelter?} $\mathtt{BOB}$ can give a partial explanation or a complete explanation. Next, we show both of them:
 
\vspace*{0.2cm}
\noindent\textbf{PARTIAL EXPLANATION}

$\mathtt{BOB}$ only uses the arguments that are part of the selected preferred extension. Thus, he answers with: $\Pa\Ea_{g_2}=\{\mathsf{A_{ep}^2,A_{ep}^3, A_{ep}^5, A_{ep}^7, A_{ep}^9, A_{ac}^2, A_{ac}^3}\}$. In natural language he would give the following answer:\textit{ $\mathsf{man\_32}$ had a fractured bone ($\mathsf{A_{ep}^2}$), the fractured bone was of his arm ($A_{ep}^3$), and it was an open fracture ($\mathsf{A_{ep}^5}$); therefore, he was severely injured ($\mathsf{A_{ep}^7, A_{ep}^9}$). Since he was severely injured I took him to the hospital ($\mathsf{A_{ac}^2, A_{ac}^3}$).}

$\mathtt{BOB}$ can also use the preferred extension of $\Aa\Fa_{ac}^{g_3}$. In this case, he gives the reasons for not sending $\mathsf{man\_32}$ to the shelter. Thus, this he answers with: $\Pa\Ea_{g_3}=\{\mathsf{A_{ep}^2, A_{ep}^3, A_{ep}^5, A_{ep}^7,A_{ep}^9, A_{ac}^4}\}$. In natural language he would give the following answer: \textit{$\mathsf{man\_32}$ had a fractured bone ($\mathsf{A_{ep}^2}$), the fractured bone was of his arm ($\mathsf{A_{ep}^3}$), and it was an open fracture ($\mathsf{A_{ep}^5}$). A fractured bone might be considered a severe injury ($\mathsf{A_{ep}^7}$), but since it it was an open fracture it was indeed a severe injury ($\mathsf{A_{ep}^9}$)}.

Notice that this explanation does not clarify completely the reasons for not sending the man to the shelter.

\vspace*{0.2cm}
\noindent\textbf{COMPLETE EXPLANATION}

In this case, $\mathtt{BOB}$ uses the sub-AFs of his individual memory records. Thus, he answers $\Ca\Ea_{g_2}=\Aa\Fa_{ac}^{g_2}=\langle \{\mathsf{A_{ep}^2,A_{ep}^5,A_{ep}^7, A_{ep}^8, A_{ep}^9,A_{ac}^2, A_{ac}^3\}, (A_{ep}^8,A_{ac}^2),}$ $\{\mathsf{(A_{ep}^7, A_{ep}^8),(A_{ep}^8, A_{ep}^7),(A_{ep}^9, A_{ep}^8)}\}\rangle$ to justify why he decided to take $\mathsf{man\_32}$ to the hospital. In natural language, this would be the answer: \textit{ $\mathsf{man\_32}$ had a fractured bone ($\mathsf{A_{ep}^2}$), the fractured bone was of his arm ($\mathsf{A_{ep}^3}$), and it was an open fracture ($\mathsf{A_{ep}^5}$). Given that he had a fractured bone, he might be considered severe injured ($\mathsf{A_{ep}^7}$); however, since such fracture was of his arm, it might not be considered a severe injure ($\mathsf{A_{ep}^8}$). Finally, I noted that it was an open fracture, which determines -- without exception -- that it was a severe injury ($\mathsf{A_{ep}^9}$). For these reasons I took him to the hospital ($\mathsf{A_{ac}^2, A_{ac}^3}$).}

The answer above answers only half the question. Let's see now the complete reason for not sending him to the shelter, which is indeed a complement of the above answer. Thus, he uses the sub-AF related to goal $g_3$: $\Ca\Ea_{g_3}=\Aa\Fa_{ac}^{g_3}=\langle \{ \mathsf{ A_{ap}^2,A_{ap}^3,A_{ap}^5, A_{ap}^7, A_{ap}^8,A_{ap}^9, A_{ac}^4 \},\{(A_{ap}^7,A_{ac}^4), }$ $\mathsf{(A_{ap}^9, A_{ac}^4), (A_{ap}^8,A_{ap}^7), (A_{ap}^7,A_{ap}^8),(A_{ap}^9,A_{ap}^8)}\}\rangle$. In natural language, this would be the answer: \textit{ $\mathsf{man\_32}$ had a fractured bone ($\mathsf{A_{ep}^2}$) and the fractured bone was of his arm ($\mathsf{A_{ep}^3}$). Given that he had a fractured bone, he might be considered severely injured ($\mathsf{A_{ep}^7}$); however, since such fracture was of his arm, it might not be considered a severe injury ($\mathsf{A_{ep}^8}$). Since the injury is not severe, the man might be sent to the shelter; however, I noted that it was an open fracture, which determines -- without exception -- that it was a severe injury ($\mathsf{A_{ep}^9}$). This last reason refutes the action of sending him to the shelter.}

Note that the complete explanation is more accurate, especially when the agent has to clarify why he did not send the man to the shelter. Note also that both complete explanations are complementary. We can say that depending on the question, the agent can use part of the entire memory goal. The agent may also use more than one memory goal in order to give satisfactory answers. 

\section{Related Work}
\label{relato}

Since XAI is a recently emerged domain in Artificial Intelligence, there are few reviews about the works in this area. In \cite{anjomshoae2019explainable}, Anjomshoae et al. make a Systematic Literature Review about goal-driven XAI, i.e., explainable agency for robots and agents. According to them, some papers propose conceptual studies and there is a lack of evaluations; almost all of the papers deal with agents that explain their behaviors to human users, and very few works tackle inter-agent explainability. 
One interesting research question was about Design, that is, the platforms and architectures that have been used to design Explainable Agency. Their results show that 22\% of the platforms and architectures have not explicitly indicate their method for generating explanations, 18\% of papers relied on \textit{ad-hoc} methods, 9\% implemented their explanations in BDI architecture. At last, other platforms and architectures used to extract explanations are: Markov Decision Process (3\%), Neural Networks (3\%), Partially Observable Markov Decision Process (3\%), Parallel-rooted-ordered Slip-stack Hierarchical Action Selection (2\%), and STRIPS (2\%).

Some works relied on the BDI model are the following. In \cite{broekens2010you} and \cite{harbers2010design}, Broekens et al. and Harbers et al., respectively, focus on generating explanations for humans about their goals and actions. They construct a tree with beliefs, goals, and actions, from which they explanations are constructed. Unlike our proposal, their explanations do not detail the progress of the goals and are not complete in the sense that do not express why an agent did not commit to a given goal. Langley et al. \cite{langley2017explainable} focus on settings in which an agent receives instructions, performs them, and then describes and explains its decisions and actions afterwards. 

Finally, Sassoon et al. \cite{sassoonexplainable} propose an approach of explainable argumentation based on argumentation schemes and argumentation-based dialogues. In this approach, an agent provides explanations to patients (human users) about their treatments. In this case, argumentation is applied in a different way than in our proposal and with other focus, they generate explanations for information seeking and persuasion.

\section{Conclusions and Future Work}
\label{conclus}

This work presented an approach for explainable agency based on argumentation theory. The chosen architecture was the BBGP model, which can be considered an extension of the BDI model. Our objective was that BBGP-based agents be able to explain their decision about the statuses of their goals, especially those goals they committed to. In order to achieve our objectives, we equipped BBGP-based agents with a structure and a mechanism to generate both partial and complete explanations. Thus, BBGP-agents not only are able to explain why their goals change their statuses but also why a goal (or goals) did not progress in the intention formation process. 

In the formalization of the BBGP model proposed in \cite{espinoza2019argumentation}, the authors also include the status \textit{cancelled}. As a future work, we intend that BBGP-based agents also generate explanations for this status. This was not done in this article because this generation goes beyond the AFs built at each stage of the intention formation process. A goal can also go back in the intention formation process. This was not taken into account and it is an interesting future research. Finally, we want to deal with more complex questions, which require more elaborate and adequate explanations. As we saw in the example, better explanations include elements of more than one AF. In this sense, a ``good'' explanation may include elements from different AFs.

\section*{Acknowledgment}
This work is fully founded by CAPES (Coordena\c{c}\~{a}o de Aperfei\c{c}oamento de Pessoal de N\'{i}vel Superior).

\bibliographystyle{unsrt}  
\bibliography{IEEEtran}  


\end{document}